\documentclass[runningheads]{llncs}
\usepackage[T1]{fontenc}
\usepackage{graphicx}
\usepackage{hyperref}
\usepackage{color}

\usepackage{xcolor}
\usepackage{amssymb}
\usepackage{multirow}
\usepackage{color, colortbl}
\usepackage{makecell}
\usepackage{gensymb}
\usepackage{mathtools}
\usepackage{algorithm,algpseudocode}
\usepackage{bm}
\usepackage[misc]{ifsym}

\usepackage{array}
\newcolumntype{C}[1]{>{\centering\arraybackslash}m{#1}}

\usepackage{capt-of}

\def\ours{\texttt{CoReEcho}}
\definecolor{TableLightCyan}{rgb}{0.88,1,1}

\begin{document}
\title{CoReEcho: Continuous Representation Learning for 2D+time Echocardiography Analysis}
\titlerunning{CoReEcho}
%
\author{Fadillah Adamsyah Maani\textsuperscript{\Letter}\inst{1} \and
Numan Saeed\inst{1} \and
Aleksandr Matsun\inst{1} \and
Mohammad Yaqub\inst{1}}
%
\authorrunning{F. Maani et al.}
%
\institute{Mohamed bin Zayed University of Artificial Intelligence, Abu Dhabi, UAE
\email{\{firstname.lastname\}@mbzuai.ac.ae}
}
\maketitle              
\begin{abstract}
Deep learning (DL) models have been advancing automatic medical image analysis on various modalities, including echocardiography, by offering a comprehensive end-to-end training pipeline. This approach enables DL models to regress ejection fraction (EF) directly from 2D+time echocardiograms, resulting in superior performance. However, the end-to-end training pipeline makes the learned representations less explainable. The representations may also fail to capture the continuous relation among echocardiogram clips, indicating the existence of spurious correlations, which can negatively affect the generalization. To mitigate this issue, we propose \ours{}, a novel training framework emphasizing continuous representations tailored for direct EF regression. Our extensive experiments demonstrate that \ours{}: 1) outperforms the current state-of-the-art (SOTA) on the largest echocardiography dataset (EchoNet-Dynamic) with MAE of 3.90 \& R2 of 82.44, and 2) provides robust and generalizable features that transfer more effectively in related downstream tasks. The code is publicly available at \url{https://github.com/BioMedIA-MBZUAI/CoReEcho}.

\end{abstract}

\section{Introduction}

Heart failure, a leading cause of mortality in clinical practice, affects 1-2\% of individuals over 40 and 10\% of those aged 60-70, representing the fastest-growing cardiac disease \cite{banegas2008heart}. Echocardiography is the preferred option for efficiently and accurately diagnosing heart conditions. From an echocardiogram video, clinicians usually assess the heart's condition by measuring biomarkers such as the ejection fraction (EF) \cite{Savarese2022,camus,ouyang2020video}, which represents the percentage of blood that can be pumped out of the left ventricle (LV). Echocardiography also allows clinicians to analyze LV characteristics to detect early myocardial infarction (MI) or heart attack \cite{Kusunose2019CJ-19-0420,degerli_early}. However, the natural characteristics of echocardiograms (e.g., speckle, operator dependence, and signal attenuation) make diagnosis challenging. Thus, recent initiatives exist to automate echocardiogram analysis through the creation of properly labeled datasets, such as \cite{camus,ouyang2020video,degerli_early}.
These publicly accessible datasets have been instrumental in advancing 2D+time echocardiography analysis, ultimately leading to a more accurate and faster diagnosis.

Early studies on EF estimation relied on the LV segmentation \cite{camus,ling:hal-03979523,lu-net}. Ouyang et al. \cite{ouyang2020video} proposed a two-stage approach in which they first segmented the LV frame-by-frame to find heartbeat cycle(s) and then used a deep learning (DL) model to regress the EF. The main limitation of these methods is the temporal inconsistency in the segmentation. These methods use 2D frames to segment the LV independently because most echocardiograms are sparsely labeled. To address this, several studies have proposed temporally-aware approaches \cite{mclas,unilvseg}
to enable video-based segmentation models for echocardiography.

Despite the promising performance of segmentation-based EF estimation, it fails on several challenging occasions where the model struggles to segment the LV properly, leading to an inability to detect any heartbeat cycle \cite{sarina_lightweight}. This challenge has led researchers to develop methods for measuring EF without relying on LV segmentation \cite{hadrien_uvt,sarina_lightweight,echocotr}. For instance, EchoCoTr \cite{echocotr} utilized a hybrid CNN-transformer video network \cite{uniformer} to estimate the EF from a sampled echocardiogram clip. This approach achieved the best EF estimation in the largest echocardiography dataset \cite{ouyang2020video}, demonstrating that the direct estimation of EF from raw video clips is more robust than the segmentation-based method.

Similar to EF estimation, there are two primary methods for classifying MI: segmentation-based and direct classification. Degerli et al. \cite{degerli_early} segmented LV walls to extract 2D+time features, followed by SVM to classify MI. Nguyen et al. \cite{nguyen2023ensemble} utilized two segmentation models to extract the features for feature ensembling followed by logistic regression, improving the MI classification. However, the improvement is inconsistent, as they perform worse when using three feature extractors. Conversely, Saeed et al. \cite{saeed_end_to_end} proposed an end-to-end method and achieved a competitive performance \cite{nguyen2023ensemble} without complex feature engineering.

\begin{figure}[t]
\centering
\begin{minipage}{0.475\textwidth}
\centering
\includegraphics[width=\textwidth]{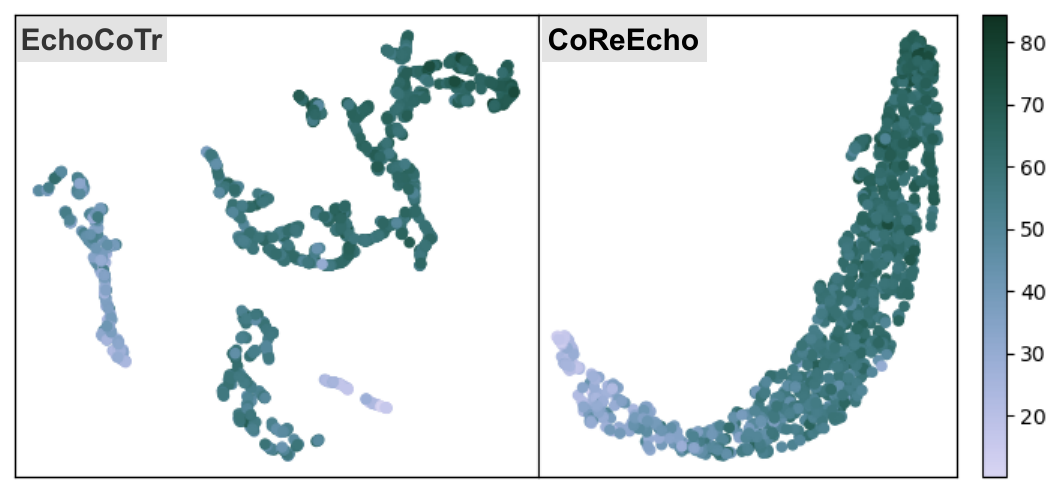}
\caption{\textbf{UMAP} of learned feature embeddings on the EchoNet \cite{ouyang2020video} test set. Unlike EchoCoTr, \ours{} assembles continuous representations in the embedding space with respect to true EF, thus mitigating clinically irrelevant relations.
}
\label{fig:umap}
\end{minipage} \hspace{0.25em}
\begin{minipage}{0.5\textwidth}
    \centering
    \begin{algorithm}[H]
    \caption{\ours{}: First Stage}\label{alg:1st}
    \scriptsize
    \begin{algorithmic}[1]
        \Require $\mathcal{F}_E$, $\mathcal{F}_R$, $\mathcal{A}$, $\mathcal{C}$, $\tau$, batch size $N$.
        \For{sampled minibatch $\{V^n,y_r^n\}_{n=1}^N$}
            \For{$n\in\{1,\dots,N\}$}
                \State Assign labels $y^{2n-1},y^{2n}=y_r^n$
                \State Sample clips $c^{2n-1}, c^{2n}$ from $\mathcal{A}(\mathcal{C}(V^n))$
                \State Extract embeddings $E^{2n-1}$, $E^{2n}$ using $\mathcal{F}_E(\cdot)$
                \State Predict $\hat{y}^{2n-1},\hat{y}^{2n}$ using $\mathcal{F}_R(\cdot)$
            \EndFor
            \State Compute $\mathcal{L}_{RnC}$ and $\mathcal{L}_{L1}$ as in Eq. \eqref{eq:l1st}
            \State Calcuate $\mathcal{L} = \mathcal{L}_{RnC} + \mathcal{L}_{L1}$
            \State update $\mathcal{F}_E$ and $\mathcal{F}_R$ to minimize $\mathcal{L}$
        \EndFor
        \State \textbf{Return} $\mathcal{F}_E$, $\mathcal{F}_R$
    \end{algorithmic}
    \end{algorithm}
\end{minipage}
\end{figure}

The aforementioned DL-based direct diagnosis methods \cite{echocotr,saeed_end_to_end} offer straightforward solutions and superior results on large-scale data yet have reduced explainability due to their end-to-end training. The resulting representations often fail to capture the underlying relations among samples \cite{zha2023rank}. In other words, the model fails to learn that hearts with similar EF values have similar features. Instead, it may learn clinically irrelevant relations, as indicated in Figure \ref{fig:umap}, that can negatively affect performance and transferability.
This highlights a critical need for methodologies that can effectively leverage the intrinsic relations among data samples. Incorporating techniques such as contrastive learning to learn from the relationships among samples can improve model performance, generalizability, and explainability \cite{supcon,pecon,zha2023rank}. For example, \cite{zha2023rank} demonstrated that modeling continuous representations boosts regression performance and transferability, an essential factor in achieving enhanced performance on small datasets \cite{saeed_end_to_end,mae,wang2023videomaev2}.

\noindent \textbf{Contributions.}
We propose \ours{}, a novel framework that enables a video analysis model to infer continuous representations when predicting a clinical outcome, achieving SOTA performance on EF regression in the largest echocardiography dataset \cite{ouyang2020video}. We further demonstrate how the \ours{}-pretrained model exhibits higher transferability than the current SOTA echocardiography model \cite{echocotr} when probing and finetuning on other small datasets. We demonstrate how our work serves as the SOTA in MI classification for a single model benchmark. Our qualitative analysis of three public datasets shows that \ours{} produces better explainable representations and visualizations than the baselines.

\section{Methodology}

\begin{figure}[t]
\centering
\includegraphics[width=0.925\textwidth]{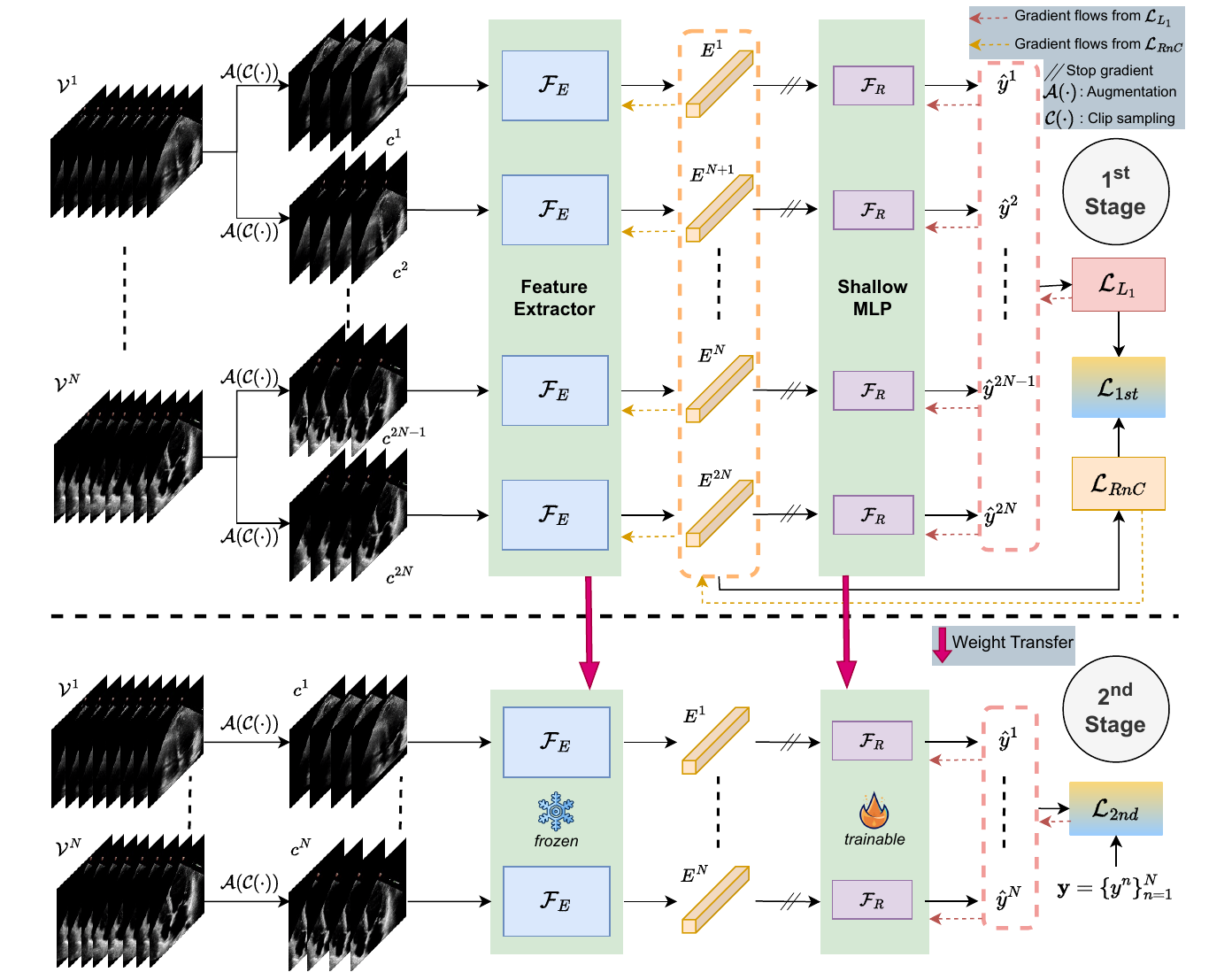}
\caption{\textbf{\ours{}}
consists of two training stages: 1) Optimize $\mathcal{F}_E$ to learn continuous relation between echocardiogram clips while simultaneously teaching $\mathcal{F}_R$ to estimate EF. 2) Perform MLP probing to further optimize $\mathcal{F}_R$ while $\mathcal{F}_E$ is frozen.}
\label{fig:method_diagram}
\end{figure}

\subsection{Problem Definition}
Given an echocardiography dataset $D\coloneqq\{V^n,y_r^n\}_{n=1}^{N_V}$ where $V^n\in \mathbb{R}^{F^n\times H \times W \times 3}$ is an echocardiogram video containing $F^n\in \mathbb{Z}^+$ frames with a frame size of $H \times W$, a corresponding EF value $y_r^n\in(0,100)$, and $N_V$ is the number of samples. We aim to train a DL-based model to predict EF value $\hat{y}^n$ from a given $V^n$. Our model is composed of two parts: a \textit{feature extractor}, $\mathcal{F}_E$, extracting a feature embedding $E^n$ from an input echocardiogram and a \textit{shallow MLP}, $\mathcal{F}_R$, predicting EF ($\hat{y}^n$) from the feature embedding $E^n$.

\noindent \textbf{Feature extractor.} Our feature extractor $\mathcal{F}_E$ is based on UniFormer-S \cite{uniformer}, driven by the fact that it achieves SOTA performance in echocardiography \cite{echocotr}. The CNNs in the first two blocks of UniFormer provide an inductive bias that helps the model learn effectively from a small dataset, followed by two global Multi-Head Relation Aggregator blocks that effectively aggregate spatiotemporal information. Thus, the embedding produced by $\mathcal{F}_E$ can be written as $E^n\in\mathbb{R}^{C_E}$.

\noindent \textbf{Shallow MLP.} We employ a shallow MLP consisting of two layers to transform the extracted features into an EF value. More specifically, $\hat{y}^n$ is given by $\hat{y}^n = W_2[g(\text{BN}_2(W_1[\text{BN}_1(E^n);1]));1]$, where BN denotes a batch norm layer \cite{bn} and $g(\cdot)$ is GELU activation \cite{gelu}. $W_1\in\mathbb{R}^{C_E\times (C_E+1)}$, and $W_2\in\mathbb{R}^{1\times (C_E+1)}$ are linear layers. Additionally, we apply dropout \cite{dropout} before every linear layer.

\noindent \textbf{Clip sampling.} DL models typically require uniform-shaped inputs for efficient processing, whereas echocardiogram videos vary in frame count. We apply clip sampling $\mathcal{C}(\cdot)$ to unify the input shapes and remove redundancy between adjacent frames before passing to $\mathcal{F}_E$. We randomly sample a clip $c^n\in \mathbb{R}^{F_C\times H \times W \times 3}$ from $V^n$ with a fixed number of frames, $F_c$, and a sampling stride of $T$.

\subsection{CoReEcho Training Pipeline}
\label{subsec:rarl}
As illustrated in Figure \ref{fig:method_diagram}, the training pipeline consists of two stages. In the \textit{first stage}, we focus on learning the continuous relationship between echocardiogram clips while teaching the model to estimate EF. We then freeze the encoder in the \textit{second stage} to refine the regression head further.

\noindent \textbf{First stage.} This stage is summarized in Alg. \ref{alg:1st}. We randomly select $N$ samples from every batch ($\{V^n,y_r^n\}_{n=1}^N$). Let $\mathcal{C}(\cdot)$ and $c$ denote the clip sampling and an echocardiogram clip, respectively. Then, we sample two clips for every echocardiogram video sample on the batch, i.e. $c^{2n-1} \triangleq \mathcal{C}(V^n)$, $c^{2n} \triangleq \mathcal{C}(V^n)$, and $y^{2n-1} \equiv y^{2n} \triangleq y_r^n$. A set of clip augmentation $\mathcal{A}(\cdot)$ is then applied to each $c^n$. Thus, the training data in each batch can be written as $\{\mathcal{A}(c^n),y^n\}_{n=1}^{2N}$. Through these processes, we expose the model to diverse augmented clips, which increases the model's robustness. Next, we extract the feature embeddings of every augmented clip, i.e., $E^n \triangleq \mathcal{F}_E(\mathcal{A}(c^n))$. Inspired by a recent work \cite{zha2023rank}, we develop a representation-aware regression loss to enable $\mathcal{F}_E$ to model the underlying continuous relation among clips in the embedding space and $\mathcal{F}_R$ to regress $\hat{y}^n$ from $E^n$. Our loss is composed of two losses, i.e., RnC loss \cite{zha2023rank} to optimize $\mathcal{F}_E$ and $L1$ loss to optimize $\mathcal{F}_R$. We apply a stop-gradient 
operator $\mathcal{SG}(\cdot)$ to prevent the gradients from $L1$ loss to $\mathcal{F}_E$ so that the regression loss does not disturb the continuous representation learning, i.e., $\hat{y}^n \triangleq \mathcal{F}_R(\mathcal{SG}(E^n))$. Let $s(A,B) \triangleq \text{exp}(-L_2(A, B)/\tau)$, $\tau\in\mathbb{R}$, and $E^l\in \mathcal{S}^{n,m}$, our loss is given by

\begin{equation}
    \label{eq:l1st}
    \mathcal{L}_{1st} = \underbrace{\frac{1}{2N(2N-1)}\sum_{n=1}^{2N}\sum_{m=1,m\neq n}^{2N} -\text{log}\frac{s(E^n, E^m)}{\sum_{E^l}s(E^n, E^l)}}_{\mathcal{L}_{RnC}}
    + \underbrace{\frac{1}{2N}\sum_{n=1}^{2N}\left|y^n-\hat{y}^n\right|}_{\mathcal{L}_{L_1}}
\end{equation}

$\mathcal{L}_{RnC}$ allows $\mathcal{F}_E$ to learn continuous representations by iteratively contrasting an anchor $n$ with a positive pair $m$ and the corresponding set of negative pairs $\mathcal{S}^{n,m}$. The negative pairs $\mathcal{S}^{n,m}$ are samples within a batch where the distance from the anchor $n$ to the negative pair is greater than the distance from the anchor $n$ to the positive pair $m$. Thus, for an $(n,m)$ pair, the corresponding negative pairs are $\mathcal{S}^{n,m} \triangleq \{E^l | l\in\{1,2,\dots,2N\}\setminus\{n\}, L_1(y^n,y^l) \geq L_1(y^n,y^m)\}$. Decreasing $\mathcal{L}_{RnC}$ implies pulling a positive pair to an anchor while pushing negative pairs away. This optimization process is analogous to a ranking mechanism, fostering continuous representations that reflect the hierarchical structure of samples in a batch with respect to the true labels. Although we only force continuity at the batch level, Markov's inequality guarantees continuity across all partitions if the $\mathcal{L}_{RnC}$ is sufficiently low and batches are sampled randomly.

\noindent \textbf{Second stage.} We continue training for several epochs further to refine $\mathcal{F}_R$ while \textit{freezing} $\mathcal{F}_E$, considering $\mathcal{F}_E$ is now capable of extracting valuable features that resemble continuous relations among samples concerning EF. Here, we only sample a clip from every video and apply the L1 loss to update $\mathcal{F}_R$:
\begin{equation}
    \mathcal{L}_{2nd}=\frac{1}{N}\sum_{n=1}^N\left|y^n-\mathcal{F}_R(\mathcal{SG}(\mathcal{F}_E(\mathcal{A}(v^n)))\right|
\end{equation}

\noindent Compared to \cite{zha2023rank}, \ours{} is more efficient since $\mathcal{F}_R$ is also optimized with additional cheap forward and backward passes in the 1st stage, requiring only a few epochs of refinement in the 2nd stage.

\subsection{Model Transferability} \label{subsec:tl}

We experiment with two prevalent transfer learning strategies, i.e., \textit{MLP probing} (MLP-P) and \textit{fine-tuning} (FT), to evaluate the transferability of a \ours{}-pre-trained model and to show how continuous latent representations can help in generalization. The probing involves attaching randomly initialized \textit{MLP} layers, $\mathcal{F}_R$, on top of a frozen pretrained \ours{}, $\mathcal{F}_E$,  to evaluate the learned latent features on a target task. In contrast, FT allows $\mathcal{F}_E$ to be updated during training for the target task, enabling $\mathcal{F}_E$ to refine its knowledge based on the nuances of the new task.

\section{Experiments}

{We used an RTX 4090 (24 GB) with PyTorch 1.12.1 and CUDA 11.6 throughout our experiments.}
We conducted experiments using three standard echocardiography benchmarks, namely EchoNet-Dynamic \cite{ouyang2020video}, CAMUS \cite{camus}, and HMC-QU \cite{degerli_early}. We tested the \ours{} training framework on \cite{ouyang2020video} to demonstrate its efficacy and then performed transfer learning experiments on \cite{camus} and \cite{degerli_early} to investigate the \ours{} pretrained model transferability. \cite{ouyang2020video} and \cite{camus} primarily deal with EF regression, whereas \cite{degerli_early} focuses on MI classification. \textit{The preprocessing, augmentation, and hyperparameter details, are provided in the Appendix}.

\noindent \textbf{EchoNet-Dynamic} \cite{ouyang2020video} is the largest publicly available echocardiography dataset containing 10,030 echocardiogram A4C videos with 112$\times$112 frame size and a varying number of frames, along with each corresponding EF reference. We follow the train-val-test split in \cite{echocotr} for fair comparisons. We sample 36 frames of each video with $T=4$ to standardize the clip shape and mitigate redundant information in adjacent frames, following \cite{echocotr}. We train our model in the 1st training stage for 25 epochs ($\tau=1.0$), followed by the 2nd stage for 5 epochs.

\noindent \textbf{CAMUS}
\cite{camus} is a 2D echocardiogram dataset containing 500 single heartbeat (SH) videos of the A2C and A4C views. It provides the EF value for each video as the main clinical metric. We only consider the A4C view since we focus on a single-view echocardiography analysis. For fair benchmarking, we perform 10-fold cross-validation (CV) following \cite{camus} splits.
Inspired by \cite{saeed_end_to_end}, we preprocess the dataset to unify the video shape. We utilize MSE loss as the cost function.

\noindent \textbf{HMC-QU} \cite{degerli_early} includes 109 A4C (SH) videos with apparent myocardial walls (72 MI and 37 normal). We follow the stratified 5-fold split provided by \cite{degerli_early,saeed_end_to_end} and strictly implement the pre-processing by \cite{saeed_end_to_end}. We also utilize the BCE loss.

\section{Results \& Discussion}

\subsection{Ejection Fraction Regression}

\begin{table}[t!]
    \centering
    {
    \begin{minipage}[t]{0.42\textwidth}
        \centering
        \caption{\textbf{EchoNet-Dynamic}: Comparison with SOTA methods on the test set. \ours{} achieves the best performance. ($^*$): \textit{reproduced performance}. SH: single heartbeat}
        \label{tab:echonet}
        \scalebox{0.85}{
        \begin{tabular}{clrrr}
            \bfseries Method & \bfseries Clip(s) & \bfseries MAE & \bfseries RMSE & \bfseries R2 \\
            \hline
            & 1 x SH & 4.22 & 5.56 & 79 \\
            \multirow{-2}{*}{\cite{ouyang2020video}} & Full video & 4.05 & 5.32 & 81 \\
            & 1 x SH & 5.32 & 7.23 & 64  \\
            \multirow{-2}{*}{\cite{hadrien_uvt}}& Full video & 5.95 & 8.38 & 52  \\
            & 1 x SH & 4.01 & 5.36 & 81 \\
            \multirow{-2}{*}{\cite{sarina_lightweight}} & Full video & 4.23 & 5.67 & 79 \\
            \cite{adacon} & 3 x Clips ($^*$) & 4.03 & 5.28 & 81 \\
            \cite{echocotr} & 3 x Clips ($^*$) & 3.98 & 5.30 & 81 \\
            \rowcolor{TableLightCyan}\ours{} & 3 x Clips & \bfseries 3.90 & \bfseries 5.13 & \bfseries 82  \\
        \end{tabular}
        }
    \end{minipage}
    \hspace{0.35em}
    \begin{minipage}[t]{0.55\textwidth}
        \centering
        \caption{Ablation study of \ours{} components. \textbf{K400}: using pretrained weight on \cite{kinetics-400}. \textbf{LP}: the 2nd stage training.}
        \label{ablation_method}
        \scalebox{0.8}{
        \begin{tabular}{C{3em}C{3em}C{3em}C{3em}r}
        \bfseries L1 & \bfseries $\mathcal{L}_{1st}$ & \bfseries K400 & \bfseries LP & \bfseries R2\\
        \hline
        \checkmark & & & & 75.11$\pm$1.73 \\
        \checkmark & \checkmark & & & 77.07$\pm$0.38 \\
        \checkmark & \checkmark & \checkmark & & 81.50$\pm$0.43 \\
        \rowcolor{TableLightCyan}\checkmark & \checkmark & \checkmark & \checkmark & \textbf{81.59}$\pm$0.18 \\
        \hline
        \multicolumn{4}{l}{EchoCoTr \cite{echocotr}} & 81.22$\pm$0.80 \\
        \end{tabular}
        }
        \vspace{0.5em}
        {
        \caption{Impact of augmentation and clip sampling. $2\times \mathcal{C}$ means we sample two clips from each video, i.e. $c^{2n-1}\neq c^{2n}$.}
        \label{ablation_method:aug_cs}
        \scalebox{0.8}{
        \begin{tabular}{C{3em}C{3em}C{3em}r}
        $\mathcal{L}_{1st}$ & $\mathcal{A}$ & $2 \times \mathcal{C}$ & \bfseries R2\\
        \hline
        \checkmark & \checkmark & \checkmark & \bfseries 77.07$\pm$0.38 \\
        \checkmark & \checkmark & & 76.93$\pm$0.44 \\
        \checkmark &  & \checkmark &  68.24$\pm$1.06 \\
        \end{tabular}
        }
        }
    \end{minipage}
    }
    {
    \caption{\textbf{CAMUS}: EF regression (A4C view) with 10-fold CV. \ours{} provides a stronger pretrained model that can transfer better on CAMUS.}
    \label{tab:result_camus}
    \scalebox{0.8}{
    \begin{tabular}{lllrrrr}
        & & & \multicolumn{2}{c}{\bfseries G\&M qual.} & \multicolumn{2}{c}{\bfseries P qual.} \\
        \multirow{-2}{*}{\bfseries Work} & \multirow{-2}{*}{\bfseries Method} & \multirow{-2}{*}{\bfseries Pretraining} & \bfseries corr & \bfseries MAE & \bfseries corr & \bfseries MAE \\
        \hline
        Wei et al. \cite{mclas} & Regression & - & 0.725 & 6.50 & \underline{0.674} & 7.53 \\
        \hline
        & Scratch & - & 0.714 & 6.20 & 0.538 & 8.28 \\
        \cline{2-7}
        & MLP & EchoCoTr \cite{echocotr} & 0.660 & 7.00 & 0.401 & 9.84 \\
        & Probing & \cellcolor{TableLightCyan}\ours{} & \cellcolor{TableLightCyan}0.730& \cellcolor{TableLightCyan}6.28 & \cellcolor{TableLightCyan}0.570 & \cellcolor{TableLightCyan}8.42 \\
        \cline{2-7}
        & Fine- & EchoCoTr \cite{echocotr} & \underline{0.799} & \underline{5.33} & 0.599 & \underline{7.49} \\
        \multirow{-5}{*}{\makecell[l]{Our \\ Implementation}} & Tuning & \cellcolor{TableLightCyan}\ours{} & \cellcolor{TableLightCyan}\bfseries0.807 & \cellcolor{TableLightCyan}\bfseries5.29 & \cellcolor{TableLightCyan}\textbf{0.693} &
        \cellcolor{TableLightCyan}\textbf{6.81}
        
    \end{tabular}
    }
    }
    \vspace{0.5em}
    {
    \caption{\textbf{HMC-QU}: MI classification (A4C view) with stratified 5-fold CV. We achieve the highest F1 score and accuracy for the single model benchmark.}\label{tab:result_hmc}
    \centering
    \scalebox{0.8}{
    \begin{tabular}{lllrrrrr}
    \bfseries Work & \bfseries Training & \bfseries Pretraining & \bfseries Sens. & \bfseries Spec. & \bfseries Prec. & \bfseries F1 & \bfseries Acc.\\
    \hline
    Degerli \textit{et al.} \cite{degerli_early} & Scratch & - & 85.97 & 70.10 & 85.52 & 85.29 & 80.24 \\
    Saeed \textit{et al.} \cite{saeed_end_to_end} & Fine-Tuning & \cite{ouyang2020video} 1-Heartbeat & 86.10 & 78.21 & 88.53 & 87.09 & 83.44 \\
    Nguyen \textit{et al.} \cite{nguyen2023ensemble} & Scratch & - & 83.30 & \bfseries{89.30} & \bfseries{94.20} & 88.30 & 85.10 \\
    \hline
    & Scratch & - & \underline{95.81} & 38.57 & 75.99 &  84.36  &  76.31\\
    \cline{2-8}
    & MLP & EchoCoTr \cite{echocotr} & 72.19 & 81.07 & 88.60 & 78.90 & 75.20 \\
    & Probing &  \cellcolor{TableLightCyan}\ours{} & \cellcolor{TableLightCyan}93.05 & \cellcolor{TableLightCyan}69.64 & \cellcolor{TableLightCyan}87.20 & \cellcolor{TableLightCyan}89.55 & \cellcolor{TableLightCyan}85.18 \\
    \cline{2-8}
    & & K400 \cite{kinetics-400} & 93.14 & 64.64 & 85.05 & 88.25 & 83.52 \\
    & & EchoCoTr \cite{echocotr} & \bfseries{98.57} & 72.50 & 88.37 & \bfseries{92.96} & \underline{89.77} \\
    \multirow{-6}{*}{\makecell[l]{Our \\ Implementation}} & \multirow{-3}{*}{\makecell{Fine-\\Tuning}} & \cellcolor{TableLightCyan}\ours{} & \cellcolor{TableLightCyan}92.95 & \cellcolor{TableLightCyan}\underline{86.07} & \cellcolor{TableLightCyan}\underline{93.55} & \cellcolor{TableLightCyan}\underline{92.91} & \cellcolor{TableLightCyan}\bfseries{90.64} \\
    \end{tabular}
    }
    }
\end{table}

\begin{figure}[t]
\centering
\includegraphics[width=0.925\textwidth]{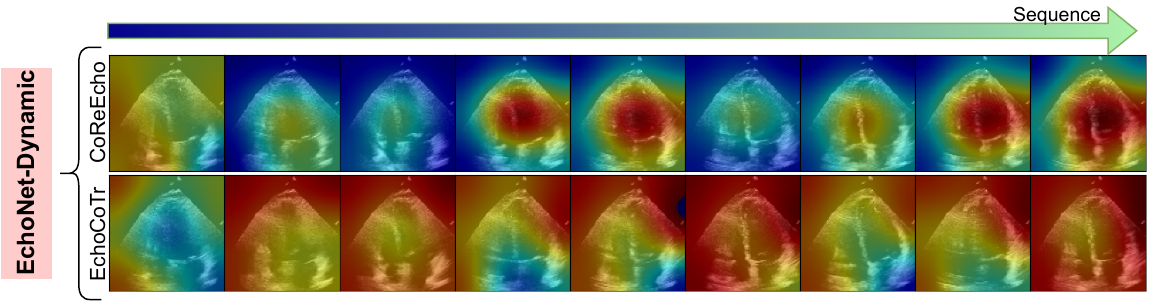}
\caption{
\textbf{Grad-CAM} \cite{grad-cam}.
\ours{} $\mathcal{F}_E$ exhibits reduced focus on backgrounds and more focus on the LV. We provide additional samples in Supplementary Materials.
}
\label{fig:cam}
\end{figure}

\noindent \textbf{Comparison with SOTA.} In Table \ref{tab:echonet}, we compare \ours{} with SOTA methods using the largest echocardiography dataset \cite{ouyang2020video}. While assessing both \ours{} and EchoCoTr \cite{echocotr}, we average the EF predictions of three sampled clips for each video to reduce the randomness introduced by the clip sampling process. \ours{} achieves the lowest MAE {of 3.90} and RMSE {of 5.13}, and the highest R2, indicating superiority over other methods. Additionally, \ours{} is more efficient than \cite{ouyang2020video,hadrien_uvt,sarina_lightweight} since it provides a direct EF regression, thus removing the need for the segmentation stage to determine a heartbeat cycle. {The training time for \ours{} is 4 hours, and the inference time is 5.95 $\pm$ 0.14 ms per clip, demonstrating our method's suitability for real-time processing.}

\noindent \textbf{Ablation study.} Table \ref{ablation_method} shows an ablation study in which we use $L_1$ loss to optimize the entire model as a baseline. We report the mean and standard deviation of the R2 scores from three runs. \textit{Using $\mathcal{L}_{1st}$ to optimize the model encoder and MLP head significantly improves the R2 score by} $+1.96\%$, suggesting that learning continuous representation in the feature embeddings leads to higher performance. Additional linear probing for five epochs (\textit{the 2nd stage}) further improves the performance, demonstrating that \ours{} is more efficient because it only requires a few epochs to refine $\mathcal{F}_R$ compared to \cite{zha2023rank}. We observe that our augmentation and clip sampling strategy are crucial in learning continuous representations, as indicated in Table \ref{ablation_method:aug_cs}.

\noindent \textbf{Qualitative analysis.} Figure \ref{fig:umap} shows the UMAP \cite{umap} of the feature embeddings across all samples in the EchoNet-Dynamic test set. \ours{} is able to represent the underlying continuous relation between the samples with respect to EF. Furthermore, Figure \ref{fig:cam} shows that compared to \cite{echocotr}, \ours{} can exhibit reduced focus on backgrounds and place a greater emphasis on the LV, which is recognized as the most informative area for assessing EF.

\subsection{Model Transferability}

\noindent \textbf{CAMUS.}
The results are presented in Table \ref{tab:result_camus}.
We calculated the performance on both Good \& Medium (\textbf{G\&M}) and Poor (\textbf{P}) quality videos following the benchmark \cite{degerli_early}. Our experiments demonstrate that in MLP probing, the \ours{}-pretrained model results in a significantly higher performance than EchoCoTr-pretrained and can even achieve an on-par performance compared to training from scratch, suggesting that\textit{\ours{} is capable of extracting robust features that can generalize well on out-of-domain (OOD) data}. Our fine-tuning experiment also shows the superiority of a model pretrained with \ours{} over EchoCoTr. Moreover, we compare our results with a direct regression method from the A4C view proposed by a SOTA method \cite{mclas} on CAMUS. We outperform the method without requiring segmentation labels for additional supervision.

\noindent \textbf{HMC-QU.} We experimented on the A4C view, and the results are presented in Table \ref{tab:result_hmc}. Similar to MLP probing on CAMUS, the \ours{}-pretrained model achieves a significantly higher F1 and accuracy than EchoCoTr-pretrained and training from scratch. This indicates that \textit{\ours{} can extract universal features that are useful for a wide range of downstream echocardiography tasks}. Allowing the encoder to be optimized (fine-tuning) results in better performance. We then compare our result with several SOTA methods on HMC-QU \cite{degerli_early,saeed_end_to_end,nguyen2023ensemble}. Our work outperforms the aforementioned methods for a single model benchmark on F1, accuracy, precision, and sensitivity.

\section{Conclusion}
We propose \ours{}, a novel training framework for a single-view echocardiography video EF estimation. \ours{} enables the encoder to represent the continuity relation across samples in the embedding space and train a shallow MLP head to leverage the extracted features to predict EF. \ours{} outperforms the state-of-the-art methods on the EchoNet-Dynamic test set. Our experiments further show that \ours{} can provide a strong pretrained model which extracts robust and universal features for single-view echocardiography downstream tasks. These observations indicate that allowing a DL-based echocardiography diagnosis model to learn more explainable representation elevates the generalization of in-domain and OOD data. Based on this, we believe \ours{} can serve as an intermediate fine-tuning task for developing an echocardiography foundation model. An intermediate fine-tuning task is critical to tailor foundation model features to be more suitable for related downstream tasks, as evidenced by several works conducted with natural \cite{wang2023videomaev2} and medical data \cite{Azizi2023}. {In addition, this continuity in representations may be used to detect distribution shifts during deployment, enabling subsequent domain adaptation to maintain performance. Exploring these topics could enhance real-world diagnostic processes, which we leave for future work.}

\begin{credits}
\subsubsection{\discintname}
The authors have no competing interests to declare that are
relevant to the content of this article.
\end{credits}

%

%
%
%
\bibliographystyle{splncs04}
\bibliography{references}

\newpage




\newcommand*{\Scale}[2][4]{\scalebox{#1}{$#2$}}%

\definecolor{LightCyan}{rgb}{0.88,1,1}
\definecolor{lightskyblue}{RGB}{225, 235, 240}
\newcommand{\muz}[1]{{\textcolor{red}{#1}}}
\newcommand{\asif}[1]{{\textcolor{orange}{#1}}}

\newcommand{\cmark}{\ding{51}}%
\newcommand{\xmark}{\ding{55}}%

\definecolor{Gray}{gray}{0.90}
\definecolor{white}{rgb}{1.0, 1.0, 1.0}

\definecolor{Lightgreen}{RGB}{218, 246, 230 }

\definecolor{label1}{rgb}{0.76,0.59,0.77}
\definecolor{label2}{rgb}{0.28,0.5,0.72}
\definecolor{label3}{rgb}{0.33,0.63,0.36}
\definecolor{label4}{rgb}{0.79,0.4,0.17}
\definecolor{label5}{rgb}{0.94,0.53,0.2}
\definecolor{label6}{rgb}{0.72,0.86,0.59}
\definecolor{label7}{rgb}{1,1,0.65}
\definecolor{label8}{rgb}{0.93,0.62,0.61}
\definecolor{label9}{rgb}{0.4,0.15,0.33}
\definecolor{label10}{rgb}{0.75,0.21,0.29}
\definecolor{label11}{rgb}{0.35,0.73,0.8}
\definecolor{label12}{rgb}{0.94,0.9,0.32}
\definecolor{label13}{rgb}{0.96,0.76,0.48}

\newsavebox{\spleen}
\savebox{\spleen}{\textcolor{label1}{\rule{1.5in}{1.5in}}}

\newsavebox{\rkid}
\savebox{\rkid}{\textcolor{label2}{\rule{1.5in}{1.5in}}}

\newsavebox{\lkid}
\savebox{\lkid}{\textcolor{label3}{\rule{1.5in}{1.5in}}}

\newsavebox{\gall}
\savebox{\gall}{\textcolor{label4}{\rule{1.5in}{1.5in}}}

\newsavebox{\eso}
\savebox{\eso}{\textcolor{label5}{\rule{1.5in}{1.5in}}}

\newsavebox{\liver}
\savebox{\liver}{\textcolor{label6}{\rule{1.5in}{1.5in}}}

\newsavebox{\sto}
\savebox{\sto}{\textcolor{label7}{\rule{1.5in}{1.5in}}}

\newsavebox{\aorta}
\savebox{\aorta}{\textcolor{label8}{\rule{1.5in}{1.5in}}}

\newsavebox{\ivc}
\savebox{\ivc}{\textcolor{label9}{\rule{1.5in}{1.5in}}}

\newsavebox{\veins}
\savebox{\veins}{\textcolor{label10}{\rule{1.5in}{1.5in}}}

\newsavebox{\panc}
\savebox{\panc}{\textcolor{label11}{\rule{1.5in}{1.5in}}}

\newsavebox{\rad}
\savebox{\rad}{\textcolor{label12}{\rule{1.5in}{1.5in}}}

\newsavebox{\lad}
\savebox{\lad}{\textcolor{label13}{\rule{1.5in}{1.5in}}}

\def\ours{\texttt{CoReEcho}}

%
\title{CoReEcho: Continuous Representation Learning for 2D+time Echocardiography Analysis}
\titlerunning{CoReEcho}

\author{Supplementary Material}
\institute{}
\maketitle 

\begin{table}[h!]
    \centering
    {
    \caption{Model input size, preprocessing steps, and augmentation implemented for each dataset. The embedding size of $C_E$ is 512. The dropout rate in the $\mathcal{F}_R$ is 0.4.}
    \scalebox{0.75}{
    \begin{tabular}{l|c|c|c}
     & \bfseries EchoNet-Dynamic & \bfseries CAMUS & \bfseries HMC-QU \\
    \hline
    \bfseries \makecell[l]{Model \\ Input Size} & $36\times112\times112$ & $16\times112\times112$ & $12\times224\times224$ \\
    \hline
    \bfseries Preprocessing & Already preprocessed & \makecell[l]{Resizing to $112 \times 112$ \\ Spline interpolation (time) \\ into 16 frames} & \makecell[l]{Center cropping to remove \\ texts \& ECG \\ Resizing to $224 \times 224$ \\ Spline interpolation (time) \\ into 12 frames} \\
    \hline
    \bfseries \makecell[l]{Augmen- \\ tation} & \makecell[l]{Zero-padding to $124\times124$ followed \\ by random cropping to $112\times112$} &  \makecell[l]{Random rotation ($\pm$20\degree) \\ Random scaling (0.8 to 1.1) \\ Random translation (10\%)} & \makecell[l]{Random rotation ($\pm$20\degree) \\ Random scaling (0.8 to 1.1) \\ Random translation (10\%)} \\
    \end{tabular}
    }
    }
    \vspace{0.5em}
    {
        \caption{Train hyperparameters. ($^*$): 25 epochs for the 1st training stage and 5 epochs for the 2nd stage.}
    \setlength{\tabcolsep}{8pt}
    \begin{tabular}{l|c|cc|cc}
     & \bfseries EchoNet & \multicolumn{2}{c|}{\bfseries CAMUS} & \multicolumn{2}{c}{\bfseries HMC-QU} \\
     & \bfseries Dynamic & MLP-P & FT & MLP-P & FT \\
     \hline
     Optimizer & \multicolumn{3}{c|}{AdamW} & \multicolumn{2}{c}{Madgrad} \\
     Batch size & 16 & 16 & 8 & \multicolumn{2}{c}{16} \\
     Epoch & 25$\cdot$5 ($^*$) & \multicolumn{2}{c|}{100} & \multicolumn{2}{c}{50} \\
     Base learning rate & 1e-4 & 2e-4 & 1e-4 & \multicolumn{2}{c}{1e-5} \\
     Optimizer momentum & \multicolumn{3}{c|}{$\beta_1$, $\beta_2$=0.9, 0.999} & \multicolumn{2}{c}{0.9} \\
     Weight decay & \multicolumn{3}{c|}{1e-4} & \multicolumn{2}{c}{0} \\
     Scheduler & Step LR & \multicolumn{2}{c|}{-} & \multicolumn{2}{c}{-}
    \end{tabular}
    \label{tab:train_hyp}
    }
    {
    \includegraphics[width=\textwidth]{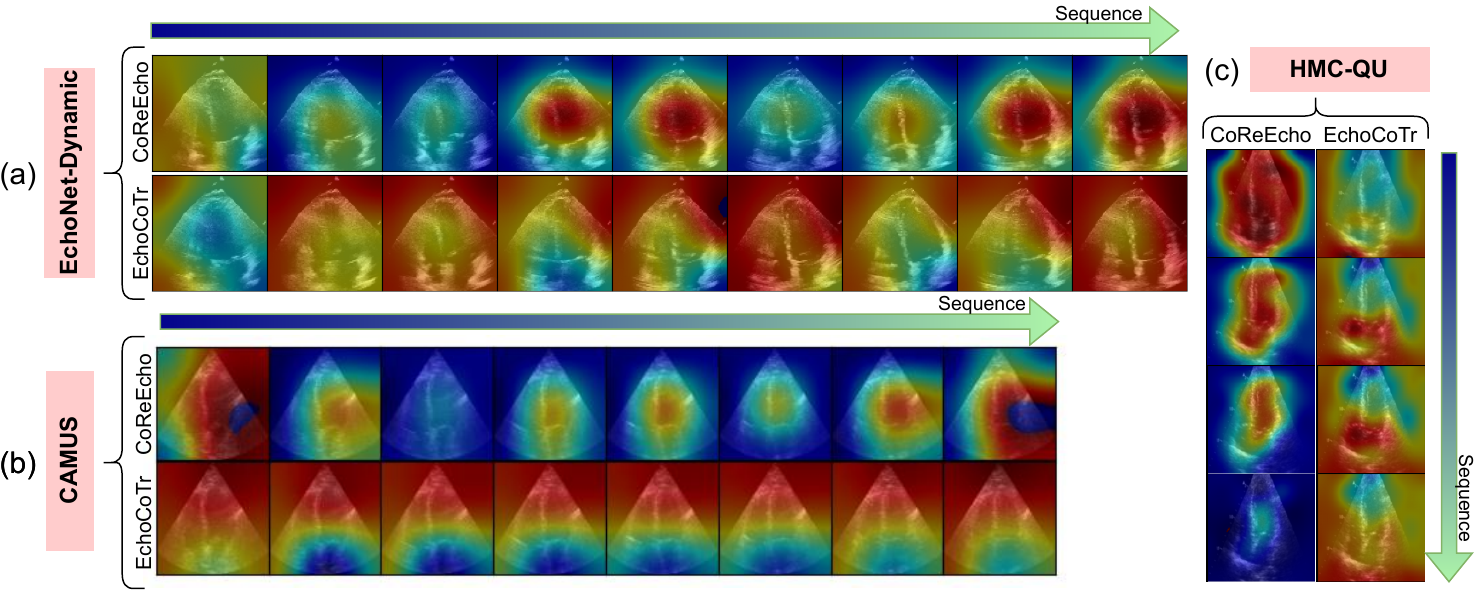}
\captionof{figure}{
\textbf{Grad-CAM}.
\ours{} $\mathcal{F}_E$ exhibits reduced focus on backgrounds and more focus on the LV.
}
\label{fig:cam}
    }
\end{table}

\begin{figure}[t]
    \centering
    \begin{minipage}{0.55\textwidth}
        \centering
        \includegraphics[width=\textwidth]{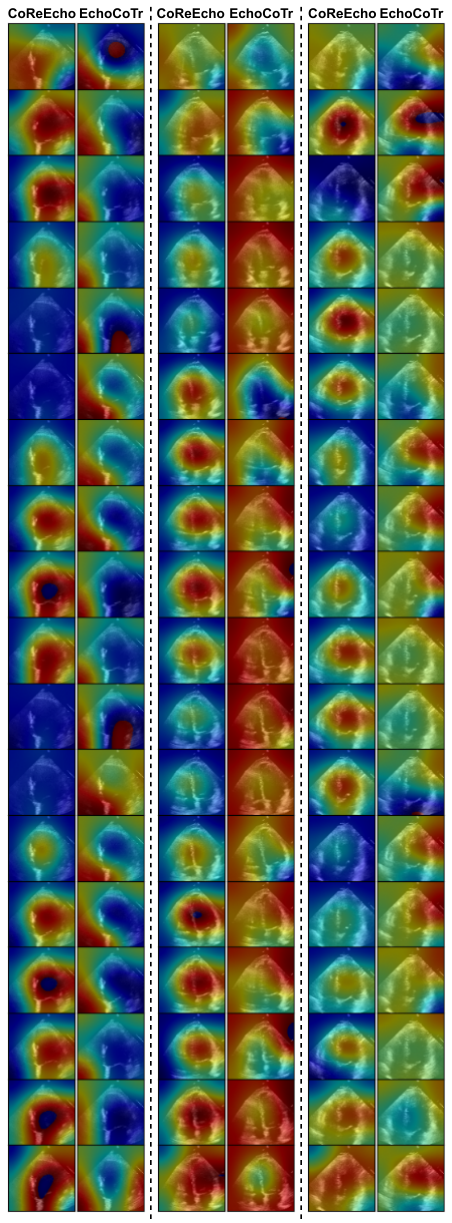}
        \caption{Grad-CAM samples on the EchoNet-Dynamic test set. \ours{} can place a higher emphasis on the LV region. \ours{} also exhibits reduced focus on backgrounds, in contrast to EchoCoTr.} \label{fig:supp_qual_echonet}
    \end{minipage}
    \begin{minipage}{0.44\textwidth}
        \centering
        \includegraphics[width=\textwidth]{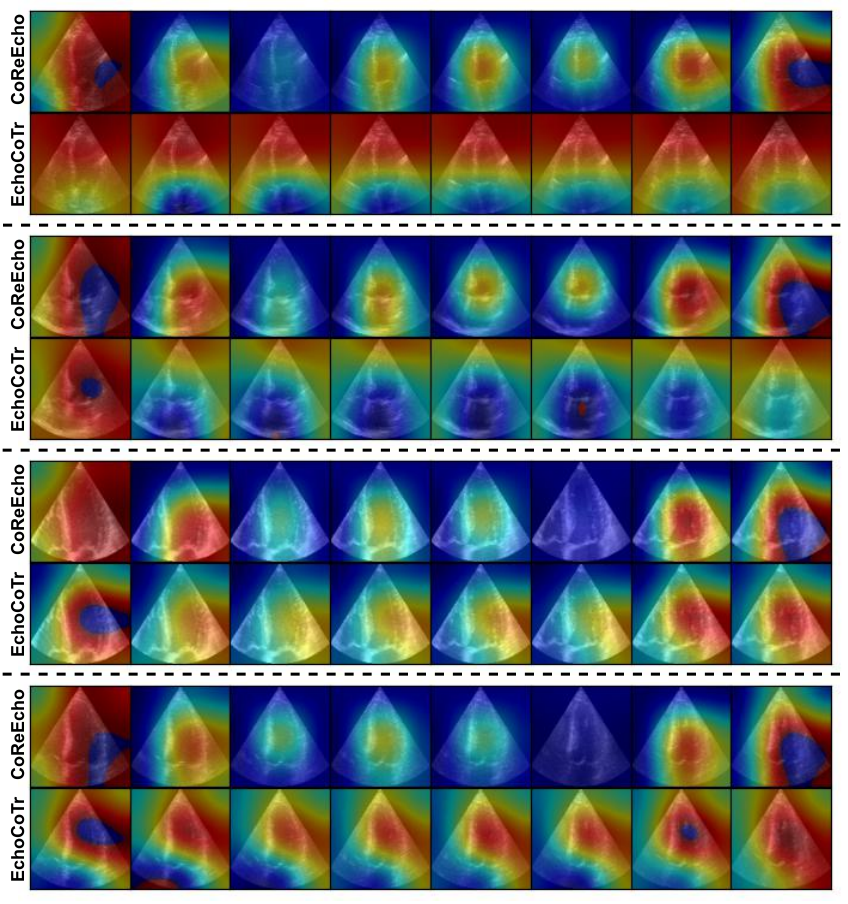}
        \caption{Grad-CAM visualization on the CAMUS dataset using the MLP probing setting.}
        \label{fig:supp_qual_camus}
        
        \hspace{1.1em}
        
        \centering
        \includegraphics[width=\textwidth]{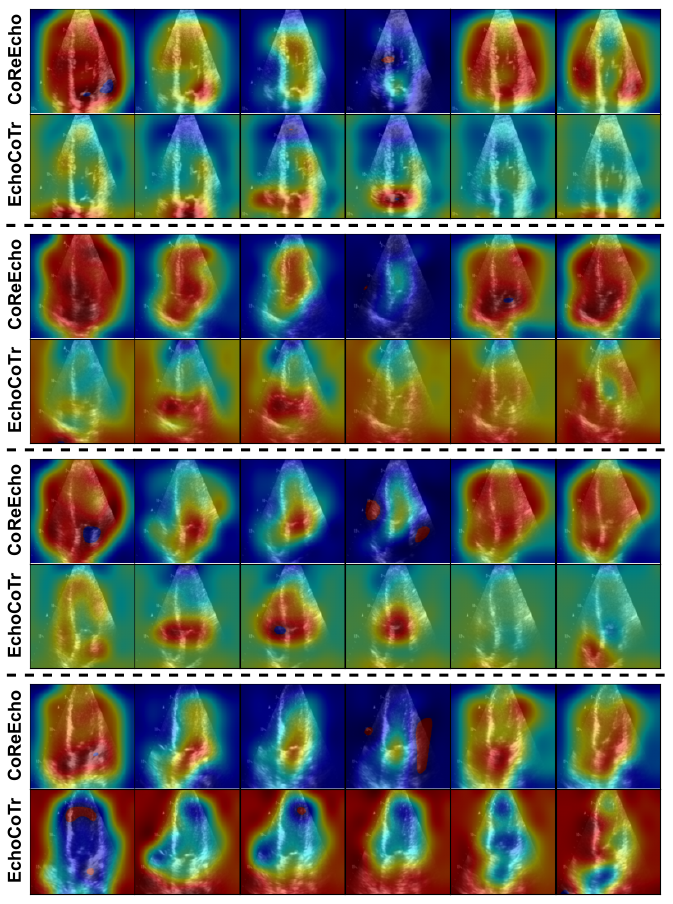}
        \caption{Grad-CAM visualization on the HMC-QU dataset using the MLP probing setting.} \label{fig:supp_qual_hmc}
    \end{minipage}
    
\end{figure}

\end{document}